# Virtual Adversarial Training for Semi-supervised Breast Mass Classification

Xuxin Chen[a], Ximin Wang[b], Ke Zhang[a], Kar-Ming Fung[c], Theresa C. Thai[d], Kathleen Moore[e], Robert S. Mannel[e], Hong Liu[a], Bin Zheng[a], Yuchen Qiu[a*]

[a] School of Electrical and Computer Engineering, University of Oklahoma, Norman, OK, USA 73019

[b] School of Information Science and Technology, ShanghaiTech University, Shanghai, China 201210

[c] Department of Pathology, University of Oklahoma Health Sciences Center, Oklahoma City, OK, USA 73104

[d] Department of Radiology, University of Oklahoma Health Sciences Center, Oklahoma City, OK, USA 73104

[e] Department of Obstetrics and Gynecology, University of Oklahoma Health Sciences Center, Oklahoma City, OK, USA 73104

## ABSTRACT

This study aims to develop a novel computer-aided diagnosis (CAD) scheme for mammographic breast mass classification using semi-supervised learning. Although supervised deep learning has achieved huge success across various medical image analysis tasks, its success relies on large amounts of high-quality annotations, which can be challenging to acquire in practice. To overcome this limitation, we propose employing a semi-supervised method, i.e., virtual adversarial training (VAT), to leverage and learn useful information underlying in unlabeled data for better classification of breast masses. Accordingly, our VAT-based models have two types of losses, namely supervised and virtual adversarial losses. The former loss acts as in supervised classification, while the latter loss aims at enhancing the model's robustness against virtual adversarial perturbation, thus improving model generalizability. To evaluate the performance of our VAT-based CAD scheme, we retrospectively assembled a total of 1024 breast mass images, with equal number of benign and malignant masses. A large CNN and a small CNN were used in this investigation, and both were trained with and without the adversarial loss. When the labeled ratios were 40% and 80%, VAT-based CNNs delivered the highest classification accuracy of 0.740±0.015 and 0.760±0.015, respectively. The experimental results suggest that the VAT-based CAD scheme can effectively utilize meaningful knowledge from unlabeled data to better classify mammographic breast mass images.

---

[*] Corresponding author: qiuyuchen@ou.edu

## 1. INTRODUCTION

Being one of the most common cancers, it is estimated that breast cancer accounts for approximately 30% of new female cancer cases across the United States in 2021[1]. In clinical settings, mammography is the standard imaging tool recommended for breast cancer screening. Interpreting mammograms by radiologists can be difficult and time-consuming particularly among dense-breasted women[2]. Meanwhile, due to expertise difference among individual clinicians, a large inter-reader variability is widely observed in clinical practice[3]. To tackle these issues, researchers have developed mammographic CAD schemes that aim at providing radiologists "a second opinion" toward better decision making[4, 5].

Currently, deep learning is the mainstream technique used by most researchers to develop advanced CAD schemes. However, most deep learning based CAD schemes are data-hungry – the performance would severely deteriorate when there is no sufficient annotated training data[6, 7]. Due to patient privacy protection and high costs, a large medical image dataset with expert annotations is often difficult to acquire, and data scarcity has become a common issue in clinical practice[8, 9]. Although annotated medical images may be scarce, unlabeled images are sometimes abundant.

Numerous studies in computer vision have demonstrated that unlabeled images usually contain valuable information, which is not present in labeled images[10, 11]. If used appropriately, unlabeled images can help achieve better performance but with fewer annotations[12]. Semi-supervised learning (SSL) is one commonly used learning paradigm to extract and exploit the information hidden in unlabeled images[13]. The SSL paradigm applies to scenarios where unlabeled images and labeled ones are relevant (i.e., in the same domain). In the recent literature of medical image analysis, there is a rising interest of applying SSL approaches to leverage unlabeled images to lessen the demand of big, annotated data. For example, in the task of breast mass classification, one study employed a semi-supervised GAN-based model to augment limited amounts of ultrasound images, and this outperformed traditional data augmentation methods[14]. In another study that focuses on segmenting breast mass from ultrasound images, an extended version of a semi-supervised temporal ensembling model was proposed to harness the underlying knowledge of unlabeled images[15], and this model yielded high segmentation accuracy with a small number of labeled images.

In this study, we adopted "virtual adversarial training" (VAT)[11] to develop a novel semi-supervised mammographic CAD scheme for breast mass classification. Different from other methods, VAT injects additive noise into the training images, aiming to enhance the robustness and generalizability of the model. VAT and VAT-inspired semi-supervised methods[16] have demonstrated great potential in the computer vision field. Nonetheless, few studies investigate the effectiveness of VAT in the medical image analysis field. Only one very recent study[17] shows that VAT is useful for identifying breast cancer from ultrasound images, which is significantly from mammogram. To the best of our knowledge, this should be one of the pilot studies to investigate the feasibility of employing VAT for the purpose of classifying breast masses from mammograms.

## 2. MATERIALS AND METHODS

**2.1 Database**

We retrospectively identified a total number of 1024 breast mass images from our existing full-field digital mammography (FFDM) image database[4, 18]. In this assembled dataset, we have 512 malignant and 512 benign cases. For each image, the mass regions were identified and resized to 128×128. The lesion types (benign or malignant) of all the cases were confirmed by biopsy examination. For this dataset, 75% of the whole dataset was used for model training and the rest 25% is for testing. Within the training dataset, a specific ratio of labeled images was used to train models. For example, if the labeled ratio is 20%, the labels of the rest 80% annotated images are hidden on purpose. We used 3 different ratios (20%, 40%, 80%), and the rest training images were considered as unlabeled data.

**2.2 Introduction of VAT**

VAT can be regarded as a more advanced type of data augmentation method, which is able to improve models' generalizability by injecting additive noise to the training data. The central idea behind the adversarial loss is to approximate a virtual adversarial direction $r_{vadv}$ such that the underlying data distribution is perturbed drastically. Suppose each input data point $x$ has the corresponding label $y$; the underlying true data distribution is denoted by $q(y|x)$, and the predicted data distribution is $p(y|x,\theta)$, where $\theta$ is the model parameters. After applying a perturbation noise $r$ onto the input data points, each perturbed input data point has a new prediction, and the predicted data distribution is denoted by $p(y|x+r,\theta)$. We want to find out the perturbation direction $r_{vadv}$ (i.e. virtual adversarial direction) that can make these two distributions as distant from each other as possible, as defined in equation (1).

$$D[q(y|x), p(y|x + r_{vadv}, \theta)] \tag{1}$$

where $r_{vadv} := arg \max_{r; \|r\|_2 \leq \varepsilon} D[q(y|x), p(y|x + r, \theta)]$, and $D[q, p]$ measures the divergence between two distributions $q$ and $p$. A weighting coefficient $\varepsilon$ was used to adjust the magnitude of the perturbation[17], which was set as 2.5 in this study.

For unlabeled images, the information of $q(y|x)$ is unknown to users, but its current estimate $p(y|x, \hat{\theta})$ provided by the trained model can be used for approximation, as shown in equation (2). This is literally equivalent to inferring "virtual" labels for the unlabeled images. The corresponding divergence value is defined as local distribution smoothness value:

$$LDS = D[q(y|x, \hat{\theta}), p(y|x + \varepsilon * r_{vadv}, \theta)] \tag{2}$$

In this investigation, the virtual adversarial direction is computationally estimated by iterative optimization[11], Based on the estimated VAT, the corresponding virtual adversarial loss $R_{vadv}(D_{ul}, \theta)$ is determined by averaging the LDS values of all the unlabeled input samples[11]:

$$R_{vadv} = \frac{1}{N} \sum_{x \in S_{ul}} LDS(x, \theta) \tag{3}$$

where $S_{ul}$ represents the set containing all the unlabeled samples, and N is the total number of the unlabeled samples. More details of the VAT method can be found in ref[11].

**2.3 VAT-based CAD scheme for mammographic breast mass classification**

Figure 1 shows the pipeline of our VAT-based CAD scheme for mammographic breast mass classification. For labeled mass images, we use the same cross entropy loss $l_{ce}$ as in supervised classification. As discussed in the above section, the virtual adversarial loss can directly work with unlabeled breast mass images, and the complete loss $l_c$ of our VAT-based model is given as follows:

$$l_c = l_{ce} + \alpha * R_{vadv}(D_{ul}, \theta) \tag{4}$$

In equation (4), the first part is a supervised loss, and the second part is the virtual adversarial loss. The importance of this adversarial loss can be adjusted by changing the value of $\alpha$. In other words, the adversarial loss of the unlabeled breast mass images serves as a regularization term to provide additional information to facilitate mass classification. By making the model robust to virtual adversarial perturbation, this regularization term can help to improve the generalizability of trained models on unseen data. We set this hyperparameter to be 1 during the experiments.

Two models were used in this study, including a large CNN and a small CNN. Both of these two models were modified from the architectures used in previous study[11]. The large CNN is composed of 19 layers, including 5 convolutional blocks, 2 max-pooling layers (MaxPool), 1 average pooling layer (AdaptiveAvgPool) and 1 linear layer. These five blocks have the same type of layers, but the filter numbers are different. Each convolutional block has 3 layers, namely, convolutional layer, batch normalization layer, and leaky rectified linear unit. The structure details of the large CNN are presented in Table 1. In the table, the output shape column provides the number of the filters (channels) and output image size. For example, the output of the third block (256, 64, 64) has a total of 256 feature maps of size 64×64. The small CNN has the same architecture as the large CNN, but only contains half of the filters. Meanwhile, the Adam optimizer was used during model training, with a learning rate of 0.001. The batch size was set to be 32. Using these two types of the losses, the CNN model generates the classification scores, indicating the likelihood that the case belongs to benign or malignant category. The model optimization and validation were performed for 3 times, and the classification accuracies were averaged for performance evaluation.

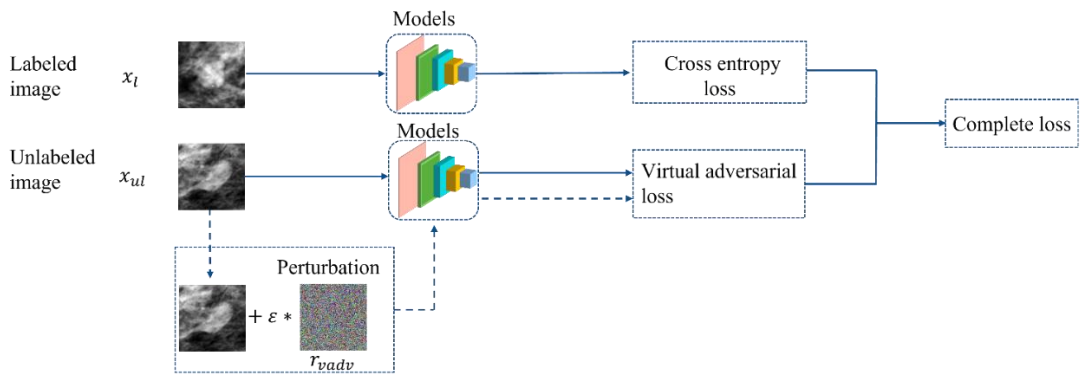

Figure 1. Pipeline of VAT-based models for mammographic breast mass classification

Table 1. Structure details of the large CNN used in this study.

| Layer Type | Output Shape |
|---|---|
| Block 1 | (128, 128, 128) |
| Block 2 | (128, 128, 128) |
| MaxPool | (128, 64, 64) |
| Block 3 | (256, 64, 64) |
| Block 4 | (256, 64, 64) |
| MaxPool | (256, 32, 32) |
| Block 5 | (128, 32, 32) |
| AdaptiveAvgPool | (128, 1, 1) |
| Linear | (2) |

## 3. RESULTS

Figure 2 demonstrates the effect of virtual adversarial perturbations on mammographic breast mass images. Based on the original image (Figure 2a), VAT generates perturbation noises (Figure 2b). Different from the random noise, the generated perturbation noise is correlated with the shape of the mass and noticeable tissues. The perturbed image (Figure 2c) is the combination of the perturbation noise and original image. The perturbed results are visually similar to the original image, but conventional CNN models may generate opposite predictions. However, our VAT-based model can help overcome this limitation, increasing the model generalizing capability on unseen data.

Table 2 shows the classification performance of two VAT-based CNN models. Models that do not use VAT are trained only on the labeled data. When 20% of the cases were labeled, the large and small CNN indicate completely different trends: adding VAT loss improves the classification performance ($0.729 \pm 0.053$ vs $0.719 \pm 0.026$) of the large CNN, but deteriorates the performance of the small CNN ($0.719 \pm 0.051$ vs $0.729 \pm 0.015$). When the labeled ratio increases to 40%, the VAT based large CNN model outperforms the model without VAT. The small CNN model, however, yielded the same performance no matter VAT is used or not. When the labeled ratio is 80%, the VAT based models consistently achieved better accuracy. Meanwhile, the highest performance of $0.760 \pm 0.015$ was yielded by the VAT based large CNN with a labeled ratio of 80%.

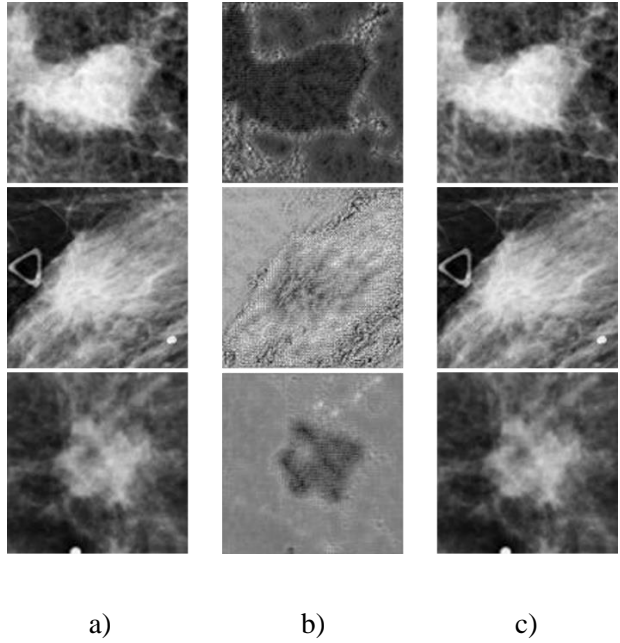

a)          b)          c)

Figure 2. a) Examples of original mass images; b) Perturbation noise generated by VAT; c) Perturbed results

Table 2. Classification accuracy of VAT-based models for semi-supervised mass classification

| **Models** | **Using VAT** | **20% labeled** | **40% labeled** | **80% labeled** |
|---|---|---|---|---|
| Large CNN | N | 0.719±0.026 | 0.719±0.051 | 0.750±0.051 |
|  | Y | 0.729±0.053 | **0.740±0.015** | **0.760±0.015** |
| Small CNN | N | **0.729±0.015** | 0.719±0.026 | 0.729±0.015 |
|  | Y | 0.719±0.051 | 0.719±0.026 | 0.740±0.078 |

# 4. DISCUSSION AND CONCLUSION

This study contributes to exploring the potential use of VAT in medical image classification tasks, via evaluating its performance in classifying benign and malignant breast masses from mammograms. We show that when the labeled ratios were 40% and 80%, VAT-based CNNs that exploit useful information from unlabeled images can outperform the CNNs trained without VAT. However, when the labeled ratio was 20%, CNNs trained without VAT achieved better performance. As discussed in **Section 2.2**, the virtual adversarial perturbations for unlabeled images are dependent on the label estimates produced by CNNs; however, when the number of labeled images becomes too low, it may be difficult for CNNs to converge and provide high-quality label estimates for unlabeled samples. Therefore, we believe that the performance deterioration is probably associated with the model's poor ability to produce reliable label estimates for unlabeled samples.

Despite the encouraging results obtained from the VAT-based models, it is still desirable to investigate 1) whether similar results can be obtained on a larger and more diversified breast mass dataset, and 2) whether changing the weight of the virtual adversarial loss can further improve the performance. In addition, this study treats the importance/ influence of each labeled or unlabeled sample equally, which may affect the performance of VAT. We expect to extend VAT by incorporating the importance of training samples in future work.

In summary, our study initially verified the feasibility of using VAT to improve the performance of classifying benign and malignant breast masses from mammograms in a semi-supervised manner. This new method may provide a new perspective regarding how to leverage unlabeled medical images into supervised tasks that have only a small number of labeled samples.

# ACKNOWLEDGEMENTS

We gratefully acknowledge the support from the research grants as follows: Oklahoma Shared Clinical & Translational Resources (OSCTR) pilot award (NIGMS U54GM104938) from the University of Oklahoma Health Sciences (OUHSC); Stephenson Cancer Center Pilot Grant, Team Grant funded by the National Cancer Institute Cancer Center Support Grant P30CA225520 awarded to the University of Oklahoma Stephenson Cancer Center.

# REFERENCES

[1]     R. L. Siegel, K. D. Miller, H. E. Fuchs *et al.*, "Cancer statistics, 2021," CA: a cancer journal for clinicians, 71(1), 7-33 (2021).

[2]     K. M. Kelly, J. Dean, W. S. Comulada *et al.*, "Breast cancer detection using automated whole breast ultrasound and mammography in radiographically dense breasts," European radiology, 20(3), 734-742 (2010).

[3]     R. P. Kruger, J. R. Townes, D. L. Hall *et al.*, "Automated radiographic diagnosis via feature extraction and classification of cardiac size and shape descriptors," IEEE Transactions on Biomedical Engineering(3), 174-186 (1972).

[4]     X. Chen, A. Zargari, A. B. Hollingsworth *et al.*, "Applying a new quantitative image analysis scheme based on global mammographic features to assist diagnosis of breast cancer," Computer methods and programs in biomedicine, 179, 104995 (2019).

[5]     X. Chen, T. C. Thai, C. C. Gunderson *et al.*, "Development of a transferring GAN based CAD scheme for breast mass classification: an initial study." 11643, 116430H.

[6]     C. Sun, A. Shrivastava, S. Singh *et al.*, "Revisiting unreasonable effectiveness of data in deep learning era." 843-852.

[7]     R. K. Samala, H.-P. Chan, L. Hadjiiski *et al.*, "Breast cancer diagnosis in digital breast tomosynthesis: effects of training sample size on multi-stage transfer learning using deep neural nets," IEEE transactions on medical imaging, 38(3), 686-696 (2018).

[8]     G. Litjens, T. Kooi, B. E. Bejnordi *et al.*, "A survey on deep learning in medical image analysis," Medical image analysis, 42, 60-88 (2017).

[9]     D. Shen, G. Wu, and H.-I. Suk, "Deep learning in medical image analysis," Annual review of biomedical engineering, 19, 221-248 (2017).

[10]    A. Tarvainen, and H. Valpola, "Mean teachers are better role models: Weight-averaged consistency targets improve semi-supervised deep learning results." 1195-1204.

[11]    T. Miyato, S.-i. Maeda, M. Koyama *et al.*, "Virtual adversarial training: a regularization method for supervised and semi-supervised learning," IEEE transactions on pattern analysis and machine intelligence, 41(8), 1979-1993 (2018).

[12]    Q. Xie, Z. Dai, E. Hovy *et al.*, "Unsupervised Data Augmentation for Consistency Training." 1-13.

[13]    X. Chen, X. Wang, K. Zhang *et al.*, "Recent advances and clinical applications of deep learning in medical image analysis," arXiv preprint arXiv:2105.13381, (2021).

[14]    T. Pang, J. H. D. Wong, W. L. Ng *et al.*, "Semi-supervised GAN-based Radiomics Model for Data Augmentation in Breast Ultrasound Mass Classification," Computer Methods and Programs in Biomedicine, 203, 106018 (2021).

[15]    X. Cao, H. Chen, Y. Li *et al.*, "Uncertainty aware temporal-ensembling model for semi-supervised abus mass segmentation," IEEE Transactions on Medical Imaging, 40(1), 431-443 (2020).

[16] D. Berthelot, N. Carlini, I. Goodfellow *et al.*, "MixMatch: A Holistic Approach to Semi-Supervised Learning." 1-11.

[17] X. Wang, H. Chen, H. Xiang *et al.*, "Deep virtual adversarial self-training with consistency regularization for semi-supervised medical image classification," Medical image analysis, 70, 102010 (2021).

[18] Y. Qiu, S. Yan, R. R. Gundreddy *et al.*, "A new approach to develop computer-aided diagnosis scheme of breast mass classification using deep learning technology," Journal of X-ray Science and Technology, 25(5), 751-763 (2017).